\def\year{2022}\relax
\documentclass[letterpaper]{article} 
\usepackage{aaai22}  
\usepackage{times}  
\usepackage{helvet}  
\usepackage{courier}  
\usepackage[hyphens]{url}  
\usepackage{graphicx} 
\usepackage{natbib}  
\usepackage{caption} 
\DeclareCaptionStyle{ruled}{labelfont=normalfont,labelsep=colon,strut=off} 
\usepackage{algorithm}
\usepackage{algorithmic}

\usepackage{newfloat}
\usepackage{listings}
\floatstyle{ruled}
\newfloat{listing}{tb}{lst}{}
\floatname{listing}{Listing}

\usepackage{lipsum}
\usepackage{amsmath}
\usepackage{comment}
\usepackage{tabularx}
\usepackage{multirow}

\newcommand{\paperTitle}{Towards the Automatic Synthesis of Interpretable Chess Tactics}

\usepackage[T1]{fontenc}

\title{\paperTitle}
\author{
   Abhijeet Krishnan \qquad
   Chris Martens
}
\affiliations{
    North Carolina State University\\

    Venture IV, 1730 Varsity Dr,\\
    Raleigh, NC 27606 \\
    akrish13@ncsu.edu, martens@csc.ncsu.edu
}

\begin{document}

\maketitle

\begin{abstract}
State-of-the-art reinforcement learning agents are capable of outperforming human experts at games like chess, Go and StarCraft II.
These agents do not simply take advantage of their digital hardware in being able to react and calculate faster than humans, but employ better strategies that lead to more victories.
Interpreting these strategies would give human players valuable insight into how to improve their play.
In this preliminary work, we propose a symbolic sub-policy model for playing chess.
Inspired by chess tactics, our model attempts to incorporate domain knowledge to improve interpretability.
We adapt patterns learned by an inductive logic programming system called PAL to derive our model.
We contribute a divergence metric to evaluate our model against a random baseline, and find a set of tactics that is able to suggest moves of similar playing strength to a human beginner.
Finally, we propose a computational evaluation scheme for the model by augmenting an off-the-shelf engine with it.
\end{abstract}

\section{Introduction}

Recent advancements in reinforcement learning (RL) have produced agents capable of competing with and even outperforming the best human experts at various games like chess~\cite{silver2018general}, Go~\cite{silver2016mastering}, Shogi~\cite{li2020suphx}, Mahjong~\cite{silver2018general}, \textit{StarCraft II}~\cite{vinyals2019grandmaster} and \textit{Dota 2}~\cite{berner2019dota}.
These agents do not simply take advantage of faster reaction and calculation abilities, but are actually employing new, better strategies that lead to more victories.
Borrowing from Jeanette Wing’s definition of computational thinking~\cite{wing2008computational}, these agents have better \textit{abstractions} than human experts for the games they’re trained to play.

Despite the existence of such agents in various competitive games, we still see human competition continue to thrive, with these agents leading to new ways of thinking and a re-evaluation of long-held beliefs about the game. 
These discoveries have, so far, involved manual or engine-assisted analysis of the games played by the agents~\cite{sadler2019game, zhou2018rethinking}.
If the agents could themselves explain their strategies and decision-making to human players, we posit that it would help improve their play.

Such chess-playing agents (chess engines) are used extensively in game analysis~\cite{smith2004modern,tukmakov2020modern} and tournament preparation~\cite{andrei-2021}.
Expert chess players utilize engine move suggestions and evaluations to analyze new lines to play~\cite{seconds}.
Most current engines use a neural network-based model with many thousands of parameters trained using deep reinforcement learning (DRL) in conjunction with a search algorithm to produce game moves.
Examples include Monte Carlo Tree Search in \textit{AlphaZero}~\cite{silver2016mastering}, Predictor+Upper Confidence Bound tree search in \textit{Leela Chess Zero}~\cite{lc0} or alpha-beta pruning in \textit{Stockfish 14}~\cite{stockfish}. 
However, this is very different to how human chess players employ pattern-recognition to produce moves~\cite{de1946het, connors2011expertise}.

Current research in the newly emerging field of explainable RL (XRL) attempts to develop methods to help humans understand RL agent decision. 
Multiple techniques like t-SNE~\cite{moore2019exploring}, trajectory clustering~\cite{osborn2018visualizing} and heatmaps~\cite{broll2019customizing} have been applied to visualize agent behaviour in games.
Symbolic policies have been investigated as interpretable representations of neural network-based policies learned via DRL.
They have been learned directly from reward signals~\cite{trivedi2021learning,landajuela2021discovering}, as surrogate models for more complex policies~\cite{verma2018programmatically}, or from input/output pairs ~\cite{derner2018data}.
However, most research in this area learns policies for optimal control in continuous environments, with discrete game environments like chess receiving little attention.

In this work, we propose a framework to learn a symbolic sub-policy model for chess.
We describe our sub-policy as being a collection of first-order logic rules which model chess tactics.
We use patterns learned by an existing inductive logic programming (ILP) system called PAL (Patterns and Learning)~\cite{morales1992first} to derive these tactics.
We contribute a divergence metric to evaluate our model of a tactic using the move evaluation capabilities of a chess engine.
We present an evaluation of a set of tactics obtained from PAL against a random baseline using our metrics.
Finally, we propose a computational evaluation of this approach by augmenting a chess engine with the synthesized tactics.
We conclude with a discussion on the limitations of this approach, along with future work.

\section{Related Work}

\subsubsection{Strategy Synthesis} A number of works attempt to learn rule-based agents using evolutionary approaches to play role-playing games like \textit{Neverwinter Nights}~\cite{spronck2004online}, board games like Checkers and Reversi~\cite{benbassat2011evolving}, cooperative games like Hanabi~\cite{canaan2018evolving}, platformers like Mario~\cite{de2018evolving}, and real-time strategy games like $\mu$RTS~\cite{marino2021programmatic}. 
Partially-applicable strategies for puzzle games have been learned using constraint satisfaction~\cite{butler2017synthesizing}.
Our model for chess tactics is learned using ILP, and incorporates domain knowledge of the concept of a tactic in order to improve interpretability.

\subsubsection{Explainable RL} Attempts to make RL agent policies amenable to human interpretation have been pursued in the XRL field.
\citet{puiutta2020explainable} provide a survey of recent XRL methods. 
An interpretability technique that has received some attention is that of training an inherently interpretable \textit{surrogate model} which matches the performance of the original agent. 
Options for this surrogate model that have been investigated include decision trees~\cite{bastani2018verifiable, coppens2019distilling, sieusahai2021explaining} and programmatic policies~\cite{verma2018programmatically,trivedi2021learning}.
Our proposed sub-policy model is only partially-applicable, and attempts to improve interpretability for chess by incorporating domain knowledge of how chess tactics are structured. 

\subsubsection{Chess Pattern Learning} Chess has been called the \textit{drosophilia}\footnote{fruit fly; easily bred and thus extensively used in genetics research} of artificial intelligence~\cite{mccarthy1990chess}. 
It has been a mainstay of AI research from the invention of the digital computer~\cite{claude1950programming} to the neural network revolution~\cite{silver2018general}. 
Given the depth of experimentation with AI techniques for chess, it is not surprising that the idea of using patterns to guide a computer to play chess is not new. 
Patterns have been used to suggest moves and guide playing strategies in middle-game positions~\cite{berliner1975representation, pitrat1977chess, wilkins1979using} and endgames~\cite{huberman1968program, bramer1977representation, bratko1982knowledge}. 
\citet{levinson1991adaptive} used weighted patterns in their Morph system as an evaluation function to guide playing strategy. 
Very recent work has attempted to directly probe neural network engines to test for the presence of human concepts~\cite{mcgrath2021acquisition}.
\citet{morales1992first} developed the PAL system to learn first-order patterns in chess using ILP. 
We build upon this work by taking advantage of modern chess engines to serve as the reference evaluation function to select learned patterns instead of hand-crafted heuristics.

\section{Background}

\subsection{Inductive Logic Programming}


Inductive logic programming (ILP) is a form of symbolic machine learning where the goal is to induce a hypothesis (a set of logical rules) that generalises given training examples~\cite{cropper2020inductive}.
It can learn human-readable hypotheses from smaller amounts of data than neural network-based models.

An ILP problem is specified by three sets of Horn clauses---$B$, the background knowledge, $E^+$, the set of positive examples of the concept, and $E^-$, the set of negative examples of the concept. 
The ILP problem is to induce a hypothesis $H$ that, in combination with the background knowledge, entails all the positive examples and none of the negative examples. Formally, this can be written as -
\begin{align*}
    &\forall e \in E^+, H \cup B \models e \text{ (i.e. }H \text{ is \textit{complete)}} \\
    &\forall e \in E^-, H \cup B \not\models e \text{ (i.e. }H \text{ is \textit{consistent)}}
\end{align*}

To make the ILP problem more concrete, we provide a toy example below.

$E^+$ and $E^-$ contain positive and negative examples of the target \verb|knight_move| relation respectively.
$B$ contains background knowledge i.e., clauses which might be useful in inducing a hypothesis for \verb|knight_move|. 

\begin{align*}
    E^+ &= \left\{\begin{array}{l}\texttt{knight\_move(d4,c6).}\\\texttt{knight\_move(d4,e6).}\\\texttt{knight\_move(d4,b5).}\\\texttt{knight\_move(d4,f5).}\end{array}\right\} \\
    E^- &= \left\{\begin{array}{l}\texttt{knight\_move(d4,d5).}\\\texttt{knight\_move(d4,b6).}\\\texttt{knight\_move(d4,e1).}\\\texttt{knight\_move(d4,h7).}\end{array}\right\} \\
    B &= \left\{\begin{array}{l}\texttt{l\_move(d4,c6).}\\\texttt{l\_move(d4,e6).}\\\texttt{l\_move(d4,b5).}\\\texttt{l\_move(d4,f5).}\end{array}\right\}
\end{align*}

From this information, we could induce a hypothesis for \verb|knight_move| as  -

\begin{align*}
    \texttt{knight\_move(From,To)} &:- \texttt{ l\_move(From,To).} \\
\end{align*}

\subsection{PAL System}

The PAL (Patterns and Learning) system was introduced in~\citet{morales1992first}. 
It attempts to use ILP to synthesize patterns for chess play, which are expressed using a subset of Horn clause logic.
It contributes a predicate vocabulary for expressing these patterns and chess positions as Horn clauses.
The pattern-learning problem is framed as an ILP problem, for which a heuristically-constrained version of the \textit{rlgg} (relative least general generalization) algorithm is used to induce plausible hypotheses.
Patterns learned can be \textit{static} and not involve any piece movement, or be \textit{dynamic} and describe multi-move tactics.
We expand upon how the PAL system formally defines and synthesizes these chess patterns.

\subsubsection{Pattern Formalism}
 
A pattern in PAL is formally defined as a non-recursive Horn clause of the form

$$
    \text{Head} \rightarrow D_1, D_2, \cdots, D_n, F_1, F_2, \cdots, F_m
$$

where,

\begin{itemize}
    \item $\text{Head}$ is the head of the pattern definition
    \item The $D_i$ are ``input'' predicates used to describe the position and represent pieces involved in the pattern
    \item The $F_j$ are instances of definitions which are either provided as background knowledge or learned by PAL, and represent the conditions (relations between pieces and places) to be satisfied by the pattern.
\end{itemize}

An example of a \textit{checking move} pattern, where a move that puts the opponent king in check is suggested, is reproduced from the paper in Figure \ref{fig:can_check}. 
A key predicate is \verb|make_move|, which determines whether a pattern is static or dynamic.
The \verb|contents| predicates are used to describe the position on the board.
The remaining predicate definitions are provided as background knowledge.

\begin{figure}
    \raggedright
    $\begin{aligned}
    \texttt{can\_check(}&\text{S1,P1,(X1,Y1),S2,king,} \\
                           &\text{(X2,Y2),(X3,Y3),Pos1)~}\leftarrow&
    \end{aligned}$
        \begin{tabular}{rl}
            &\verb|contents|(S1,P1,(X1,Y1),Pos1), \\
            &\verb|contents|(S2,king,(X2,Y2),Pos1), \\
            &\verb|other_side|(S1,S2), \\
            &$\neg$~\verb|in_check|(S2,(X2,Y2),P1,(X1,Y1),Pos1), \\
            &\verb|make_move|(S1,P1,(X1,Y1),(X3,Y3),Pos1,Pos2), \\
            &\verb|in_check|(S2,(X2,Y2),P1,(X3,Y3),Pos2).
        \end{tabular}
    \caption{PAL rule for the \texttt{can\_check} pattern. A piece (P1) can check the opponent's King after moving to (X3,Y3).}
    \label{fig:can_check}
\end{figure}

\subsubsection{Pattern Synthesis}

The input to the PAL generalization algorithm is a set of pattern definitions (both predefined and learned) along with a description of a chess position (as ground unit clauses).
The algorithm extends \citeauthor{buntine1988generalized}'s method for constructing the \textit{rlgg} of two clauses to multiple clauses.
It uses the following constraints and heuristics to limit hypothesis size and increase the algorithm's generalisation steps -

\begin{itemize}
    \item Disallowing variables in the head or body of a rule which are not \textit{connected} to a literal i.e., not equal to a variable of that literal
    \item Labeling constants occurring in the ground literals of a rule body to make patterns piece-invariant
    \item Restricting the legal moves from a position to only be those which introduce a new predicate name or remove an existing predicate name
\end{itemize}

PAL uses an automatic example generator to manually guide the generalization algorithm towards learning desired concepts.
Given an example of the target concept, the generator \textit{perturbs} the example to create a new example for which a classification label must be provided.
To restrict the example space searched, the automatic example generator attempts to generate examples which specialize the current hypothesis in case of a prior positive example, or generalize it in case of a prior negative example.
We refer interested readers to the original thesis for further details.

\section{Methodology}

\subsection{Chess Tactic Model}

We conceptualize our sub-policy model as a \textit{chess tactic}.
Formally, we define a tactic as a first-order logic rule that can bind to a chess position.
A position is expressed in first-order logic using an appropriate predicate vocabulary. 
If a tactic binds to (matches) a particular position, it suggests a move (or moves) to be played.
The moves suggested must be legal in the given position.
This is described in Figure \ref{fig:tactic-defn} as a Prolog pseudo-definition.

\begin{figure}
    \raggedright
    \verb|tactic|(Position)~$\leftarrow$
        \begin{tabular}{rl}
            &\verb|matches|(Position), \\
            &!, \\
            &\verb|suggested|(Move1,Move2,$\cdots$,MoveN), \\
            &\verb|legal|(Position,Move1), \\
            &\verb|legal|(Position,Move2), \\
            &$\vdots$ \\
            &\verb|legal|(Position,MoveN).\\
        \end{tabular}
    \caption{A Prolog pseudo-definition for a tactic. ``!'' is the Prolog cut operator.}
    \label{fig:tactic-defn}
\end{figure}

A single tactic, or even a set of tactics, does not represent complete policy for playing chess.
This is because we might encounter a position for which no tactic matches.
In this case, our model cannot make a move.
There might also be positions to which multiple tactics apply, in which case an arbitration process for selecting a single move among the various suggestions is not obvious.

\subsection{Tactic Utility Metrics}

We introduce two metrics --- \textit{coverage} and \textit{divergence}, to measure the utility of a learned tactic.

\subsubsection{Coverage} A tactic $t$'s coverage for a set of positions $P$ is calculated as ---

$$
    \text{Coverage}_t = \frac{|P_\text{match}|}{|P|}
$$

where a position $p \in P_\text{match}$ if there is a binding assignment of the variables in the rule head of the tactic $t$ to the position $p$.

\subsubsection{Divergence} To measure the quality of moves suggested by a tactic, we extend a metric previously used to analyse world chess champions~\cite{guid2006computer,guid2011using,romero2019computer} to multiple moves using \textit{discounted cumulative gain} (DCG)~\cite{jarvelin2002cumulated}.
A move's \textit{error} in a position $p$ is measured by comparing it to the best move suggested by the engine in that position. 
This comparison is done quantitatively by using the engine's move evaluation function $eval(\cdot, p)$. 
In case an engine evaluates a position to be a `Mate in X' rather than a centipawn score, we assign an arbitrary large value to the evaluation.

\begin{align*}
    \text{Error}(move,p) = |eval(move_{\text{engine}},p) - eval(move,p)|
\end{align*}

Since a tactic might suggest multiple moves, we propose the use of DCG as a metric to compare ranked move suggestion lists. 
Assuming the list of suggestions output by a tactic to be in ranked order, we obtain a list of best moves from the engine of similar length as the suggestions, and compare the two using DCG.
Thus, the final divergence metric for a tactic $t$ over a set of positions $P$ is ---

\begin{align*}
    \text{Divergence}_{t} = 
    \frac{1}{|P_\text{match}|}\sum_{p \in P_\text{match}}\sum_{i=1}^{|M_t|}\frac{\text{Error}(m_i,p)}{\log_2{(1+i)}}
\end{align*}

where $M_t$ is the ranked list of move suggestions output by a tactic, and $m_i$ is the $i^{\text{th}}$ move in $M_t$. divergence of a tactic (to the reference engine) is low and close to 0 when its suggestions are similar in evaluation to the engine's best moves, and takes on large values when it differs significantly.

\subsection{Implementation using PAL}

We use the PAL system to synthesize tactics.
We select seven patterns that PAL was shown to learn, and modify them to output a move suggestion.
These patterns and their verbal definitions are listed in Table \ref{tab:patterns}. 
All patterns learned other than \verb|pin| are 1-ply \textit{dynamic} patterns, which means they include a single \verb|make_move| predicate in the rule body looking ahead one move.
We modify these patterns to introduce a \verb|suggestion| predicate with the same variables as \verb|make_move|. For \verb|pin|, which is a static pattern as learned by PAL, we convert it into a dynamic pattern as shown in Figure \ref{fig:pin} and introduce the \verb|suggestion| predicate in the same way.

\begin{figure}[ht]
    \raggedright
    $\begin{aligned}
    \texttt{pin(}&\text{S1,P1,(X1,Y1),S2,king,} \\
                 &\text{(X2,Y2),S2,P3,(X3,Y3),(X4,Y4),Pos1)~}\leftarrow&
    \end{aligned}$
        \begin{tabular}{rl}
            &\verb|sliding piece|(P1,(X1,Y1),Pos1), \\
            &\verb|make_move|(S1,P1,(X1,Y1),(X4,Y4),Pos1,Pos2) \\
            &\verb|sliding piece|(P1,(X4,Y4),Pos2), \\
            &\verb|stale|(S2,P3,(X3,Y3),Pos2), \\
            &\verb|threat|(S1,P1,(X4,Y4),S2,P3,(X3,Y3),Pos2), \\
            &{$\begin{aligned}
            \texttt{in\_line(}&\text{S2,king,(X2,Y2),S2,P3,} \\
            &\text{(X3,Y3),S1,P1,(X4,Y4),Pos2).} \\
            \end{aligned}$} \\
        \end{tabular}
    \caption{Modified PAL rule for the \texttt{pin} pattern to convert it into a tactic}
    \label{fig:pin}
\end{figure}

\begin{table*}[ht]
    \centering
    \begin{tabularx}{\textwidth}{l|X}
        \multicolumn{1}{c}{\textbf{Pattern}} & \multicolumn{1}{|c}{\textbf{Definition}} \\
        \hline
        \verb|can_threat| & A piece (P1) can threaten another piece (P2) after making a move to (X3,Y3) \\
        \verb|can_fork| & A piece (P1) can produce a fork to the opponent's King and piece (P3) after making a move to (X4,Y4) \\
        \verb|can_check| & A piece (P1) can check the opponent's King after a moving to (X3,Y3) \\
        \verb|discovered_check| & A check by piece (P2) can be ``discovered'' after moving another piece (P1) to (X4,Y4) \\
        \verb|discovered_threat| & A piece (P1) can threaten an opponent's piece (P3) after moving another piece (P2) to (X4,Y4) \\
        \verb|skewer| & A King in check by a piece (P1) ``exposes'' another piece (P3) when it is moved out of check to (X4,Y4) \\
        \verb|pin| & A piece (P3) cannot move because it will produce a check on its own side by piece (P1) \\
    \end{tabularx}
    \caption{Patterns learned by the PAL system that are used to create tactics}
    \label{tab:patterns}
\end{table*}

\section{Evaluation}

We wish to investigate whether the synthesized tactics tend to suggest good moves to play.
We do this by measuring coverage and divergence for each of our tactics over a set of positions using both a strong and a weak reference engine.
For our strong reference engine, we use \textit{Stockfish 14}, the winner of the TCEC 2020 Championship~\cite{haworth202120}.
For our weak reference engine, we use Maia Chess~\cite{mcilroyyoung2020maia}, a chess engine trained to produce human-like moves. 
We use the \verb|maia1| model, which is targeted toward 1100 ELO (a measure of relative playing strength, and roughly equal to a beginner).
We limit the search depth to 1-ply for both Stockfish 14 and Maia 1100 to resemble our tactics.
As a baseline, we use a random tactic which is applicable to all positions and produces a random legal move in the position.
We limit the number of suggestions from a tactic to 3, and assume the output order of tactic suggestions as the intended ranked order.
For ease of implementation, we manually translated the tactics from Prolog definitions to Python functions.
We use 5000 games from the January 2013 archive of standard rated games played on lichess.com~\cite{lichess-db}.
For each game, we generate positions by iterating through the move list, making the move, and adding the resulting position to the evaluation set.
In total, we generate 325,830 positions.

\section{Results and Analysis}

We summarize the results of our evaluation in Table \ref{tab:results}.

From the high coverage values obtained, we conclude that tactics like \verb|can_threat| and \verb|discovered_threat| are too general, whereas tactics like \verb|discovered_check| are too specific.
Tactics like \verb|can_check|, \verb|can_fork| and \verb|skewer| strike a balance between these extremes.

From the divergence metrics calculated using Maia-1100 (our weak engine), we see that most of our tactics have lower divergence scores than our random baseline, indicating that they tend to produce moves which are evaluated somewhat similarly to a weak engine's best moves.
For Stockfish 14, however, all our tactics have higher divergence scores than random, indicating that they do not tend to produce moves similar to a strong engine.
Thus, we qualitatively conclude that our tactics resemble that of a beginner chess player.

\begin{table}[ht]
\begin{tabular}{|l|r|rr|}
\hline
\multicolumn{1}{|c|}{\multirow{2}{*}{Tactic}} & \multicolumn{1}{c|}{\multirow{2}{*}{Coverage}} & \multicolumn{2}{c|}{Divergence}                                        \\ \cline{3-4} 
\multicolumn{1}{|c|}{}                        & \multicolumn{1}{c|}{}                          & \multicolumn{1}{l|}{SF14} & \multicolumn{1}{l|}{Maia} \\ \hline
\verb|can_threat|                                   & 0.96                                           & \multicolumn{1}{r|}{378.94}    & 9.22                           \\
\verb|can_check|                                    & 0.45                                           & \multicolumn{1}{r|}{549.19}    & 4.02                           \\
\verb|can_fork|                                     & 0.32                                           & \multicolumn{1}{r|}{676.45}    & 4.67                           \\
\verb|discovered_check|                            & $\approx$0                                              & \multicolumn{1}{r|}{338.55}    & 18.64                          \\
\verb|discovered_threat|                            & 0.96                                           & \multicolumn{1}{r|}{375.97}    & 1.19                           \\
\verb|skewer|                                        & 0.22                                           & \multicolumn{1}{r|}{748.4}     & 5.41                           \\
\verb|pin|                                           & 0.79                                           & \multicolumn{1}{r|}{526.45}    & 4.9                            \\
\verb|random|                                        & \textbf{1}                                              & \multicolumn{1}{r|}{\textbf{328.09}}    & \textbf{8.28}                           \\ \hline
\end{tabular}
\caption{Coverage and DCG for each tactic}
\label{tab:results}
\end{table}

\section{Proposed Evaluation}

We propose an experiment to investigate whether the identified set of tactics are useful for a human player to learn in order to generate good moves.
To do this, we will measure the win-rate of a chess engine against a version of itself augmented with these tactics.
As a human proxy, we plan to use Maia Chess, specifically \verb|maia1| model targeted toward 1100 ELO.
We will play games between the engines of 30 minutes + 5 seconds time control, following the TCEC League rules~\cite{tcec} and starting from the default start position.
As before, we will limit the evaluation depth of the search tree for the augmented and unaugmented engines to 1. 
We measure the divergence scores for our tactics with a strong reference engine (Stockfish 14).
We will modify the action-selection procedure of Maia Chess to utilize the first suggestion of the lowest divergence tactic applicable to a given position, instead of the engine's move choice. 
This is made explicit in Algorithm \ref{alg:augmented}. 
Finally, we will compare the win rates of the augmented engine against the unmodified version of itself over multiple games.
Our hypothesis is that the augmented engine will have a significantly higher win-rate, enabling us to conclude that the set of tactics tend to suggest good moves.

\begin{algorithm}[ht]
    \caption{Augmented engine move selection}
    \label{alg:augmented}
    \textbf{Input}: set of tactics $T$, position $p$, chess engine move selection procedure $C.$\verb|make_move|$(\cdot)$\\
    \textbf{Output}: legal move in position $p$
    \begin{algorithmic}[1] 
        \STATE move $\leftarrow$ $C.$\verb|make_move|$(p)$
        \STATE min\_divg $\leftarrow \infty$ 
        \FOR{$t \in T$}
        \IF{$t$ matches $p$ \AND divg$(t) <$ min\_divg}
        \STATE move $\leftarrow$ $t.$\verb|suggestion|
        \STATE min\_divg $\leftarrow$ divg$(t)$
        \ENDIF
        \ENDFOR
        \STATE \textbf{return} move
    \end{algorithmic}
\end{algorithm}

\section{Conclusion and Future Work}

We have described a symbolic sub-policy model for chess inspired by the pattern-action model of chess tactics.
We have used patterns learned by an ILP system to construct these tactics.
We have contributed a metric for measuring the divergence of these tactics to a reference chess-playing agent.
We evaluated a set of tactics learned by a chess pattern learning system using our metric to find that they resembled a weak engine, but were not similar to a strong one.

We use patterns learned by PAL to obtain our tactics.
However, PAL uses manual labeling of generated examples to learn specific concepts, and requires additional effort to convert the learned patterns into tactics for our model.
We aim to investigate the automatic learning of tactics from a dataset of chess positions and move suggestions using ILP as implemented by modern systems like Popper~\cite{cropper2021learning}.
Future work could investigate alternate ILP algorithms to use our divergence metric as a loss function to optimize tactics for~\cite{evans2018learning}.

\begin{figure}[h]
    \centering
    \includegraphics[width=0.9\columnwidth]{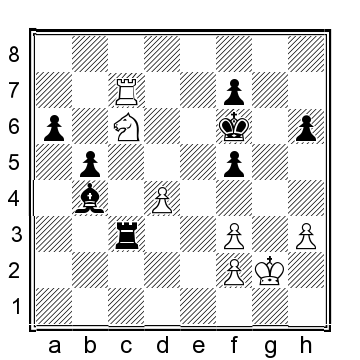}
    \caption{An example of the limitations of our 1-ply \texttt{can\_fork} tactic. White has no immediate forking move here, leading to the tactic not matching. However, if they play \textbf{1.~Nxb4}, then Black's best response is \textbf{1.~...~Rxc7} which allows a fork with \textbf{2.~Nd5+} leading to the capture of the rook.}
    \label{fig:fork-alt}
\end{figure}

Our tactic model is loosely inspired by how chess tactics are learned and practiced.
However, our tactics are limited to looking 1-\textit{ply} in the future i.e., they can only recognize the presence of a matching pattern in the immediately next position. 
Many chess tactics suggest \textit{combinations} of moves, a series of moves where the matching pattern shows up only in a particular sequence (see Figure \ref{fig:fork-alt}). 
Extending the tactic model to express and recognize such combinations will be a useful avenue for future work.
We also wish to investigate the expression of longer-term plans from chess literature like centre control and pawn structure using tactics.

Our appeal to the interpretability of these tactics rests on similar claims made regarding the interpretability of rule-based strategies. 
Future work will involve rigorously testing these assumptions with user studies using evidence-based measures of interpretability \cite{lage2019human,kliegr2021review}.
Specifically, we wish to investigate the ease of learning and applying these tactics in real games played by human players.



\bibliography{main}

\end{document}